\documentclass{svproc}
\usepackage{url}
\usepackage{amsmath}
\usepackage{amssymb}
\usepackage{graphicx}
\usepackage{subcaption}
\usepackage{array}
\usepackage{comment}

\begin{document}
\mainmatter              
\title{Autonomous overtaking trajectory optimization using reinforcement learning and opponent pose estimation}

\titlerunning{Autonomous overtaking}  
%
\author{Matej Rene Cihlar\inst{*} \and Luka Šiktar \and Branimir Ćaran \and Marko Švaco}

%
\authorrunning{Cihlar et al.} 

\tocauthor{Matej Rene Cihlar, Luka Šiktar, Branimir Ćaran, Marko Švaco}
%
\institute{University of Zagreb, Zagreb 10000, Croatia,\\
Faculty of mechanical engineering and naval architecture,\\
\email{matej.cihlar@fsb.unizg.hr}
}

\maketitle              

\begin{abstract}
Vehicle overtaking is one of the most complex driving maneuvers for autonomous vehicles. To achieve optimal autonomous 
overtaking, driving systems rely on multiple sensors that enable safe trajectory optimization and overtaking efficiency. 
This paper presents a reinforcement learning mechanism for multi-agent autonomous racing environments, enabling overtaking 
trajectory optimization, based on LiDAR and depth image data. The developed reinforcement learning agent uses pre-generated 
raceline data and sensor inputs to compute the steering angle and linear velocity for optimal overtaking. The system uses LiDAR 
with a 2D detection algorithm and a depth camera with YOLO-based object detection to identify the vehicle to be overtaken and its 
pose. The LiDAR and the depth camera detection data are fused using a UKF for improved opponent pose estimation and trajectory 
optimization for overtaking in racing scenarios. The results show that the proposed algorithm successfully performs 
overtaking maneuvers in both simulation and real-world experiments, with pose estimation RMSE of (0.0816, 0.0531) m in (x, y).

\keywords{reinforcement learning, autonomous racing, sensor fusion, LiDAR, depth camera, YOLO}
\end{abstract}
\section{Introduction}

Autonomous racing has emerged as a challenging domain for testing self-driving algorithms and autonomous overtaking methods. 
Driving at the dynamic limits in adversarial conditions requires advanced vehicle control and perception algorithms. The 
autonomous racing environment is highly stochastic and complex, making it an ideal testbed for pose estimation and reinforcement 
learning (RL) algorithms. Additionally, introducing opponents into the environment makes the
problem more challenging in terms of safety and performance. This paper focuses on optimizing overtaking maneuvers in a 
multi-agent racing environment using a reinforcement learning approach, with opponent pose estimation required for optimal 
overtaking. Estimating the pose of the opponent, as opposed to obstacle avoidance or treating the opponent as a static object, 
results in fewer crashes and more optimal driving. To produce a robust estimate of the opponent pose, an unscented Kalman 
filter (UKF) is implemented to fuse data from a LiDAR and a depth camera. The LiDAR pose estimation algorithm produces a 2D 
position (x, y) of the opponent vehicle from the 2D LiDAR scan, while the depth camera uses YOLO to detect the opponent vehicle's 
bounding box, calculate its depth, and enhance the pose estimation. Using YOLOv8 object detection, specifically fine-tuned for 
the opponent vehicle, the distance to the opponent is calculated. The yaw angle of the opponent is not estimated because it can be 
directly determined from the vehicle kinematic model.  

For opponent pose estimation, which is used to calculate optimal driving and overtaking trajectories, this paper employs sensor 
fusion with a UKF \cite{ukf2000}. A similar sensor fusion approach has been applied in autonomous driving by Hwang et al. 
\cite{Hwang2024ROS2IO}, but not in autonomous racing scenarios on the F1TENTH platform. The proposed method combines 2D LiDAR 
pose estimation \cite{Zhang-2017-26536} and depth camera data for opponent pose estimation using YOLOv8 object detection models 
\cite{yolo2016}.
Current state-of-the-art RL algorithms for autonomous racing are designed to follow optimal racing lines in single-agent racing 
environments.
Ghignone et al. \cite{ghignone2024TCDriverAT} presented a deep RL algorithm that follows optimal racing lines using a 
trajectory-conditioned reinforcement learning approach. Budai et al. \cite{budai2025endtoend} presented a deep RL algorithm 
that uses a curriculum learning approach to learn optimal racing lines in a single-agent environment. Despite these advances, 
most RL algorithms are simulation-based and lack real-world validation \cite{Betz_2022}. This paper focuses on an adversarial 
real-world multi-agent racing environment where overtaking maneuvers are incorporated into the reward formulation.

Evans et al. \cite{Evans2023HighSpeedAR} proposed a system in which the car learns overtaking maneuvers from a predefined expert 
dataset, validated on an F1TENTH scaled racing platform, similar to the most current RL algorithm implementations 
\cite{f1tenth2020_paper}. Chisari et al. \cite{chisari2021learning} presented trained RL models that can be transferred to 
real-world scenarios using a policy output regularization approach and a lifted action space for smooth actions in race car 
driving.

The main contribution of this paper is the combination of RL overtaking maneuvers that use opponent pose, estimated using UKF 
sensor fusion, validated on a real-world F1TENTH racing platform.

\section{Methodology}
\begin{figure} [h!]
    \centering 
    \begin{subfigure}[b]{0.45\linewidth}
        \centering
        \includegraphics[width=1.12\linewidth]{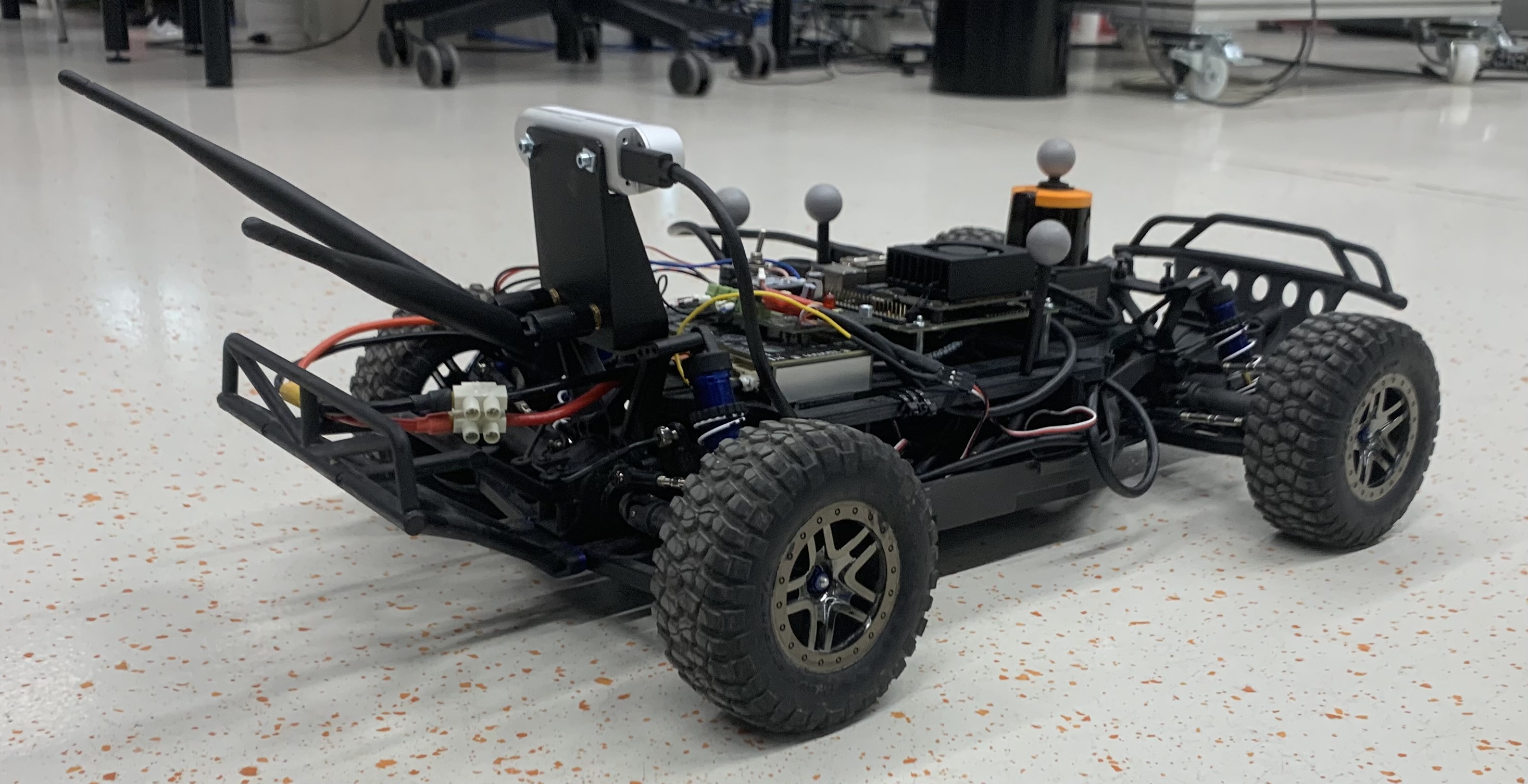}
    \end{subfigure}
    \hfill
    \begin{subfigure}[b]{0.45\linewidth}
       \includegraphics[width=0.95\linewidth]{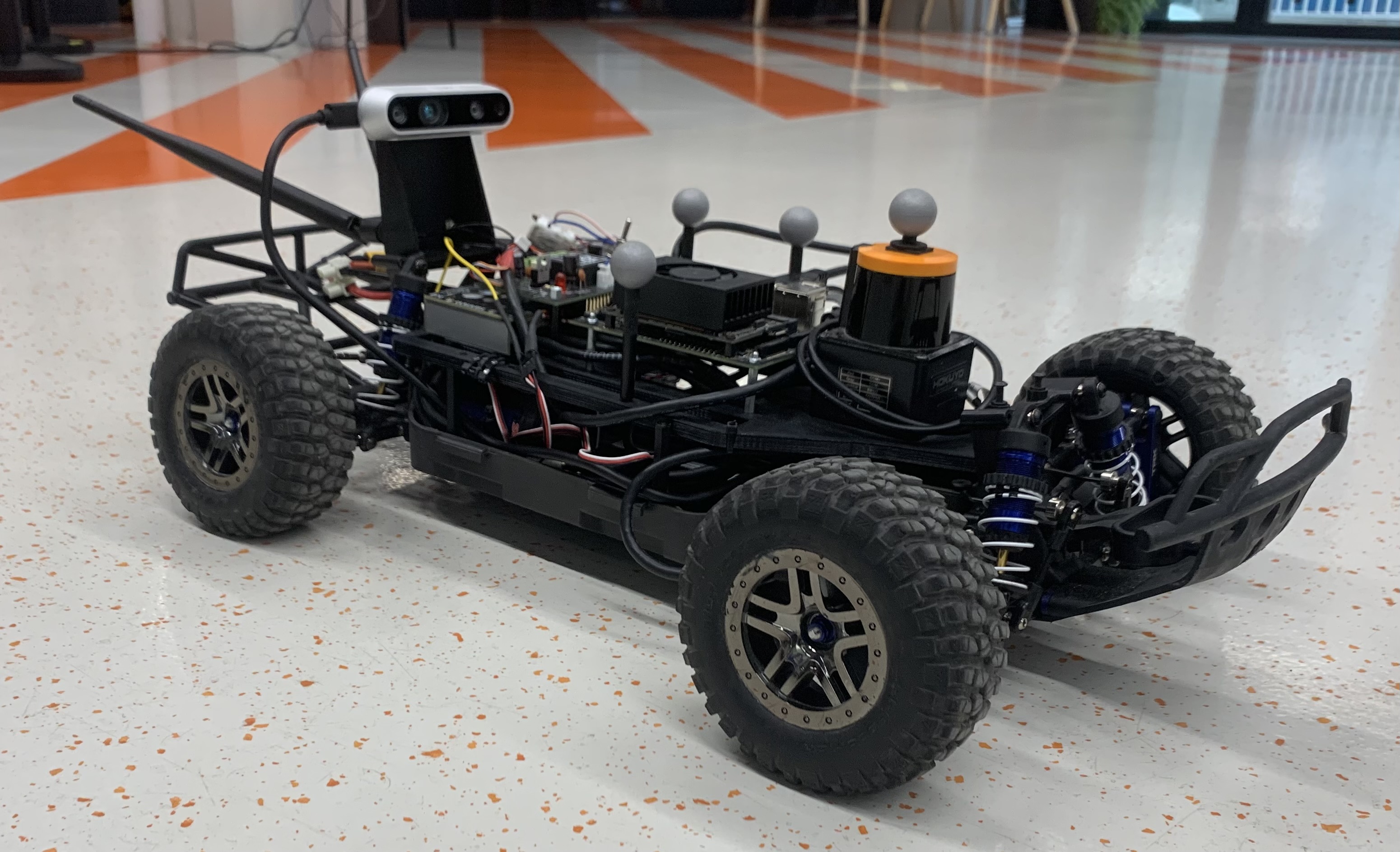} 
    \end{subfigure}
    
    \caption{\label{fig:f1tenth_setup}F1TENTH racing platform setup}
\end{figure}

This paper consists of two main components: pose estimation using sensor fusion and a reinforcement learning algorithm for 
overtaking maneuvers. The physical platform used for real-world testing is based on the F1TENTH platform \cite{f1tenth2020_paper}. 
For processing, a Jetson Xavier NX computer is used, while a Hokuyo 2D LiDAR and an Intel RealSense D435 depth camera provide sensor 
inputs. The final setup is shown in Figure \ref{fig:f1tenth_setup}.

\subsection{Opponent pose estimation}
Opponent pose estimation, relative to the RL agent, is performed using an onboard 2D LiDAR with an object detection algorithm 
and an Intel RealSense D435 depth camera with RGB-based neural network object detection and depth data. The pose estimates are fused 
using an Unscented Kalman Filter (UKF). The overall system performance is validated in real-world scenarios using the F1TENTH 
racing platform ground truth position, obtained with the Optitrack motion capture system.
\\
For LiDAR-based opponent pose estimation, an object detection algorithm from \cite{lidar_estimation_code} uses clustering and 
rectangle fitting. The most likely opponent car cluster is then selected using the UKF estimate from the previous timestep.
\\
The depth camera uses both RGB and depth data for opponent pose estimation. For the RGB data, the YOLOv8 
\cite{yolo2016} object detection model is used to detect the opponent car's bounding box. The 
model is trained on a custom dataset of 120 images using Roboflow \cite{gallagher2025useyolov12}, and since only one F1TENTH 
car is available, an Astro robot is used as the opponent. This should not affect the algorithm 
behavior since the model can be trained on any car as long as it can detect it reliably.  
\\
The distance estimation is done by taking the height of the bounding box and applying the general reciprocal function:

\begin{equation} 
    distance = \frac{a}{bbox\_height + b} + c,
\end{equation}

where the parameters were determined experimentally. This approach is valid since the opponent is always seem from the side, 
so the height changes only with distance, not with orientation.
The yaw angle is not estimated because it can be approximated from the Ackermann steering model, which limits 
the direction of travel 
to the direction of orientation. 
\\
From the bounding box center pixel location, the angle around the z-axis of the opponent relative to the agent 
can be calculated, and depth data can be extracted. Since the 
depth distance actually measures the back of the opponent car, an average offset was calculated from the 
Astro robot dimensions, and added to the depth distance. These measurements were then passed to the UKF.
\\
For the sensor fusion, an Unscented Kalman Filter (UKF) is implemented based on \cite{ukf2000}. The UKF 
state vector is defined as $[x, y, v_x, v_y]$ and a constant velocity model is used for 
the process. Since LiDAR and depth camera measurements are in different formats and from different 
sensors, two different measurement models are defined, including the static transformations for the 
sensor frames that are measured manually. Since sensors have different publishing rates, the UKF is 
updated on every mesurement. In this step the LiDAR 
cluster is determined. 

\subsection{Reinforcement Learning Algorithm}
Reinforcement learning (RL), as a learning paradigm in which the agent optimizes its behavior through 
interactions with the environment, uses a reward system that optimizes the agent's behavior for a specific goal. Defining the 
environment, the agent's action space,
and the reward structure, the agent is precisely trained for the goal-directed behavior. In this paper, a multi-agent RL algorithm 
(MARL) is 
implemented to learn overtaking maneuvers in a racing environment, which means that the agent learns using the experience from both 
cars.
\\
For learning, the Proximal Policy Optimization (PPO) algorithm \cite{schulman2017ppo} is used, 
which improves training stability by using a specific clipped loss function:
 
\begin{equation}
    L^{CLIP}(\theta) = \hat{\mathbb{E}}_t[\min(r_t(\theta)\hat{A}_t,\text{clip}(r_t(\theta),
    1-\epsilon,1+\epsilon)\hat{A}_t)],
\end{equation}

where $r_t(\theta)$ is the probability ratio between new and old policy, $\hat{A}_t$ is the
advantage estimate at time step t and $\epsilon$ is a clipping range hyperparameter. A custom implementation is used based on \cite{tabor_youtube_code}. 
The neural network consists of a fully connected feedforward network with two hidden layers, each with 256 neurons and ReLU 
activation functions. The input layer is of size 162 (observation size), and the output 
layer is of size 2 (action size) with a tanh activation function. For the multi-agent training 
setup, opponent’s actions are also generated by the same NN policy and the data from both agents is 
used for training.
\\
The F1TENTH gym environment \cite{f1tenth2020_paper} is used in this paper. It simulates the 
F1TENTH car as a simple bicycle model. It also models collisions using a 2D occupancy grid 
map and simple boxes as car representations. The environment simulates a 2D LiDAR sensor, equivalent to the 
real car's Hokuyo LiDAR, with 1080 points per scan along 270 degrees. The F1TENTH gym simulation environment enables the LiDAR 
data and the car odometry data extraction for RL. The environment also allows up to two cars for multi-agent racing 
scenarios. 
\\
The observation space consists of agent LiDAR data, agent and opponent car odometry and their future waypoints. 
Since agent and opponent data are explicitly defined, it reduces the necessity of using the full LiDAR scan, so for 
simplicity every 10th LiDAR point is taken, resulting in 108 points, which describe the
surrounding obstacles. The waypoints are defined as 10 future waypoints and are taken 
also for the opponent, so the agent could theoretically predict the opponent's future trajectory 
and plan its overtaking maneuver accordingly. Both waypoint sets are taken from a 
precalculated racing line that is generated from the map files using the minimum curvature 
approach \cite{schulman_CL2Raceline}. The observation is defined in the local reference frame for better generalization.
\\ 
The action space is defined as the continuous steering angle and the linear velocity commands, derived as the output of the 
RL neural network. The steering 
angle is scaled to $\pm$ 0.34 rad, which is around 20 degrees, and the linear velocity is scaled 
from 0 to 3 m/s, which are the max steering angle and the max safe velocity of the F1TENTH car in real-world scenarios.
\\
To promote the learning of successful overtaking maneuvers and optimal racing lines, the reward 
signal is the weighted sum of seven components: the velocity reward, the progress reward, the overtaking 
reward, the raceline penalty, the collision penalty, the heading penalty and the smoothness penalty. The velocity 
reward is defined as the current linear velocity of the agent. The progress reward is 
defined as the progress along the raceline, the smoothness penalty is defined 
as the change in steering angle between two consecutive time steps. The raceline and heading penalties 
are implemented as the distance and the yaw angle errors from the raceline. The overtaking reward is 
1 when the agent is in front and -1 when the opponent is in front, when the agents are close 
(less than three waypoints apart) the overtaking reward was scaled linearly between -1 and 1.

\section{Results}

\begin{figure} [h!]
\centering
\includegraphics[width=0.96\linewidth]{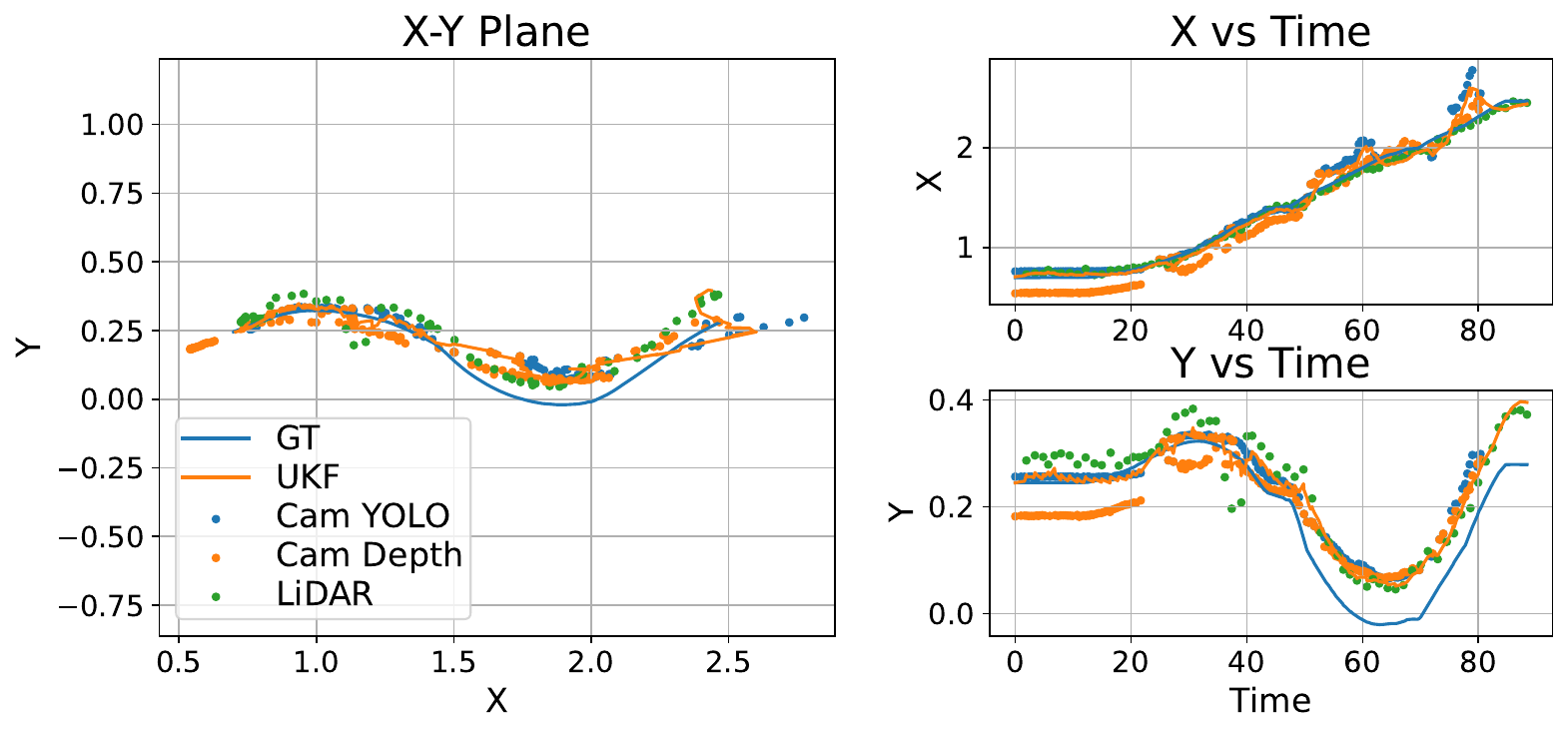}
\caption{\label{fig:UKF_results}UKF opponent pose estimation results}
\end{figure}

The opponent pose estimation successfully tracks the opponent pose, as shown in Figure 
\ref{fig:UKF_results}. The UKF estimates follow the ground truth closely for 
position, but the estimation error increases with distance. The RMSE error values are shown in Table 
\ref{tab:RMSE_values}. The UKF performs better than the individual sensors in the y direction, but is worse 
than the LiDAR in the x direction. The UKF allows for more robust tracking because the LiDAR has a wider FOV, while 
the camera knows 
which LiDAR cluster is the opponent. Without the camera, the LiDAR can start tracking the wrong 
object. Furthermore, the UKF can run at 10 Hz and give continuous results from the 1-2 Hz sensor updates, which
are needed for the continuous overtaking trajectory optimization. Due to the limitations in the implementation (slow 
sensor update rates) the algorithm is evaluated at lower speeds to prioritize estimation precision over reaction time. This can be improved with better hardware and more optimized sensor algorithm implementations.

\begin{table} [h!]
\centering
\caption{\label{tab:RMSE_values}RMSE values for the opponent pose estimation}
\begin{tabular}{|c|c|c|c|c|}
 \hline
 \rule{0pt}{3ex}Sensor & LiDAR & Depth & YOLO & UKF \\
 \hline
 \rule{0pt}{3ex}RMSE x [m] & 0.0430 & 0.1164 & 0.1106 & 0.0816 \\
 \hline
 \rule{0pt}{3ex}RMSE y [m] & 0.0652 & 0.0628 & 0.0575 & 0.0531 \\
 \hline
\end{tabular}
\end{table}

\begin{figure}[h!]
    \centering
    
    \begin{subfigure}[b]{0.3\linewidth}
        \centering
        \includegraphics[width=\linewidth]{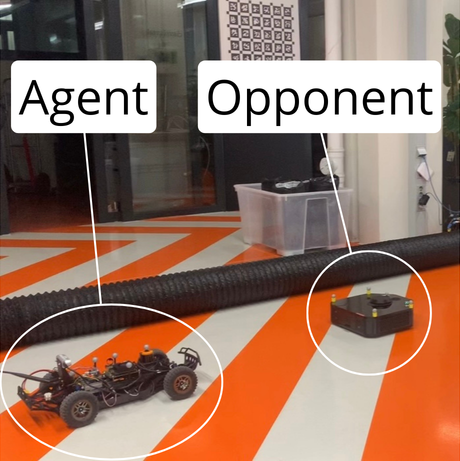}
        \caption{Starting position}
        \label{fig:MARL1_real}
    \end{subfigure}
    \hfill
    \begin{subfigure}[b]{0.3\linewidth}
        \centering
        \includegraphics[width=\linewidth]{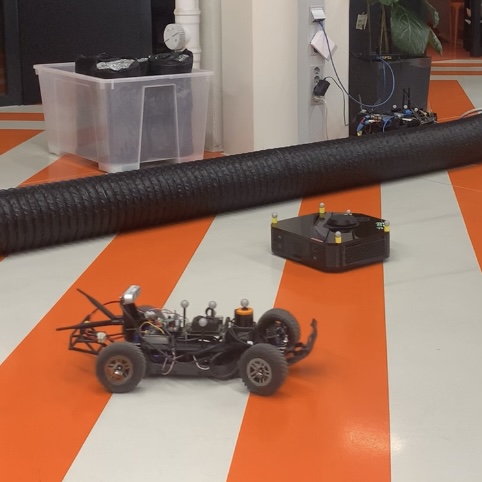}
        \caption{Position 2}
        \label{fig:SMARL3_real}
    \end{subfigure}
    \hfill   
    \begin{subfigure}[b]{0.3\linewidth}
        \centering
        \includegraphics[width=\linewidth]{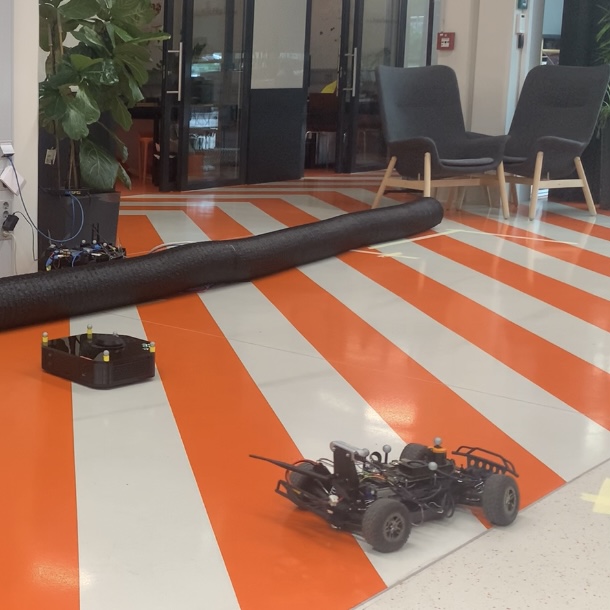}
        \caption{Position 3}
        \label{fig:MARL5_real}
    \end{subfigure}
    
    \begin{subfigure}[b]{0.3\linewidth}
        \centering
        \includegraphics[width=\linewidth]{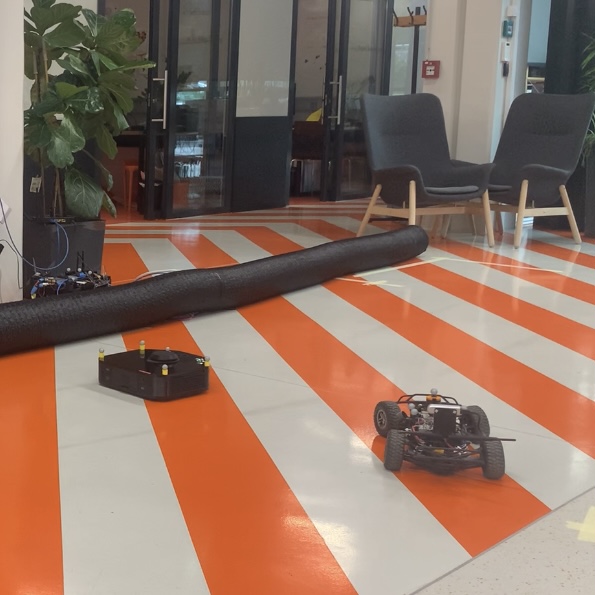}
        \caption{Position 4}
        \label{fig:MARL7_real}
    \end{subfigure}
    \hfill   
    \begin{subfigure}[b]{0.3\linewidth}
        \centering
        \includegraphics[width=\linewidth]{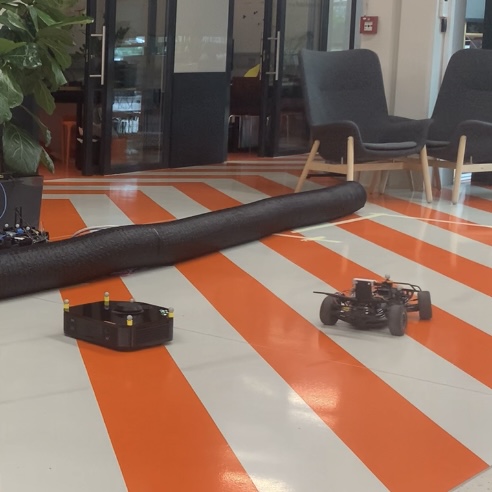}
        \caption{Position 5}
        \label{fig:MARL9_real}
    \end{subfigure}
    \hfill
    \begin{subfigure}[b]{0.3\linewidth}
        \centering
        \includegraphics[width=\linewidth]{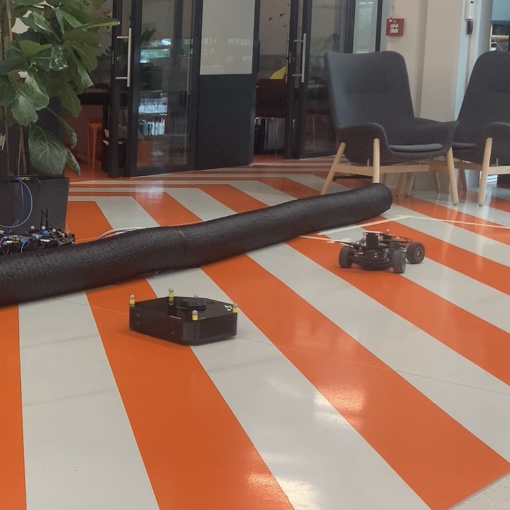}
        \caption{Final position}
        \label{fig:MAR11_real}
    \end{subfigure}
    
    \caption{MARL real-world overtaking maneuver}
    \label{fig:MARL_real_pictures}

\end{figure}

\begin{figure} [h!]
\centering
\includegraphics[width=1\linewidth]{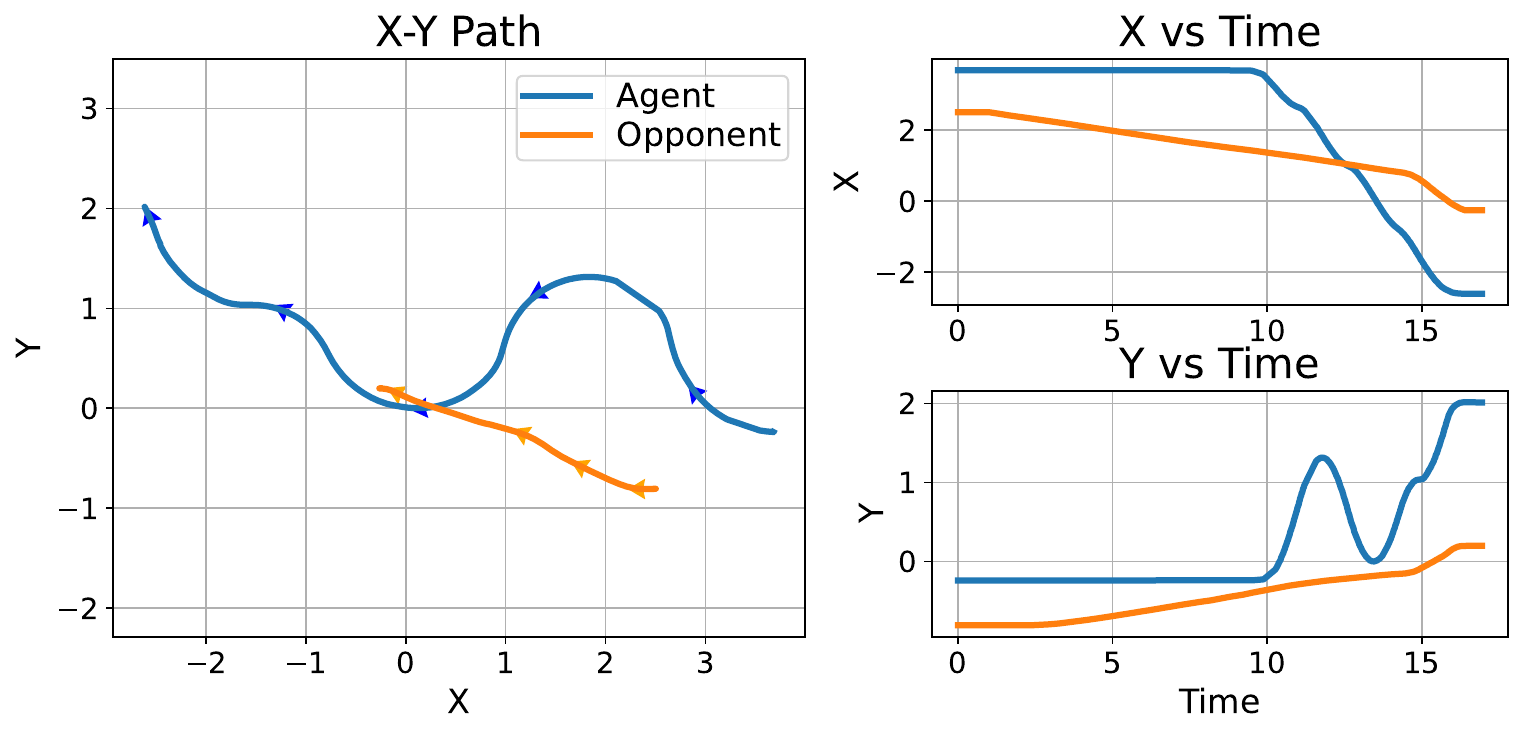}
\caption{\label{fig:MARL_real}MARL real world overtaking maneuver results}
\end{figure}

The final RL algorithm is deployed on the real F1TENTH car and tested in an overtaking scenario. The 
agent successfully overtook the opponent in multiple runs without collisions. The final trajectories of 
the opponent and the agent from one of the tests are shown in Figure \ref{fig:MARL_real}. The trajectory 
has oscillations, which are present in the simulation but are amplified in the real-world conditions due to 
noise and delays. Figure \ref{fig:MARL_real_pictures} shows one 
of the tests in multiple frames through time. The testing setup is a corridor with a similar width to 
the simulated racing track. Since the agent is trained on a large fixed-width track, the agent is not 
able to generalize to the smaller looped track real-world scenario. However, the agent is 
still able to overtake the opponent successfully when the real-world conditions are similar to the 
training conditions.

\section{Conclusion}

Autonomous racing in multi-agent environments is a complex task due to the dynamic nature and overtaking  
maneuvers. This paper presented a reinforcement learning algorithm using PPO that focuses on overtaking 
optimality. Opponent pose estimation is successfully implemented using sensor fusion of 2D LiDAR and 
depth camera data using a UKF. The pose estimate is validated using a motion capture system and the final 
RMSE values are 
presented. The final RL algorithm is successfully deployed on a real F1TENTH racing 
platform. Future work will focus on improving trajectory smoothness and increasing pose estimation update rates.

\section*{Appendix}

\url{https://github.com/CRTA-Lab/roborace_stack.git}
%
%

\bibliographystyle{spmpsci_unsrt.bst}
\nocite{*}
\bibliography{references.bib}

\end{document}